\pdfoutput=1

\documentclass[11pt]{article}

\usepackage{ACL2023}

\usepackage{times}
\usepackage{latexsym}

\usepackage[T1]{fontenc}

\usepackage[utf8]{inputenc}

\usepackage{microtype}

\usepackage{tabularx}
\usepackage{enumitem}
\usepackage{inconsolata}
\usepackage{graphicx}
\usepackage{bm}
\usepackage{bbm}
\usepackage{multirow}
\usepackage{xcolor}
\usepackage{colortbl}
\usepackage{float}
\usepackage{tikz}

\newcommand*\circled[1]{\tikz[baseline=(char.base)]{
            \node[blue,shape=circle,draw,inner sep=1pt] (char) {#1};}}

\title{OPI at SemEval 2023 Task 1: Image-Text Embeddings and Multimodal Information Retrieval for Visual Word Sense Disambiguation}

\author{Sławomir Dadas\\
  National Information Processing Institute, Warsaw, Poland\\
  \texttt{sdadas@opi.org.pl}}

\begin{document}
\maketitle
\begin{abstract}
The goal of visual word sense disambiguation is to find the image that best matches the provided description of the word's meaning. It is a challenging problem, requiring approaches that combine language and image understanding. In this paper, we present our submission to SemEval 2023 visual word sense disambiguation shared task. The proposed system integrates multimodal embeddings, learning to rank methods, and knowledge-based approaches. We build a classifier based on the CLIP model, whose results are enriched with additional information retrieved from Wikipedia and lexical databases. Our solution was ranked third in the multilingual task and won in the Persian track, one of the three language subtasks.
\end{abstract}

\section{Introduction}

Visual word sense disambiguation (VWSD) is a task in the field of multimodal natural language processing, in which the goal is to identify the intended meaning of a target word in a given context by selecting the most appropriate image from a set of candidate images. Finding images corresponding to the correct meaning of the word might improve the performance of methods combining text and visual information such as image search engines, visual question answering, or image generation models.

SemEval 2023 workshop hosted a task on visual word sense disambiguation. The task involved selecting the best matching image out of ten candidates given a short textual description. The descriptions usually consisted of two words: the target word and the context word \citep{raganato-etal-2023-semeval}. For example, the phrase \emph{andromeda tree} contains the ambiguous target word \emph{andromeda} and the context word \emph{tree}, which indicates a specific meaning of the target word. The task organizers provided three datasets, of which the trial and training datasets contained phrases in English, while the test dataset was multilingual and consisted of English, Italian, and Persian subsets. Participants were allowed to submit their solutions for a particular language or for all three languages. The systems were ranked according to the average accuracy score from three language-specific subtasks.

In this paper, we describe our system for the VWSD shared task. The backbone of our solution is a classifier using multimodal CLIP embeddings \citep{radford2021learning, cherti2022reproducible}, which has been enriched with features extracted from Wikipedia and dictionaries. These knowledge sources are used to retrieve textual and image data, providing additional information useful for determining the correct meaning of the target word. Our system was ranked third in the multilingual task and took first place in the Persian subtask. The source code of our system is publicly available, as well as the fine-tuned models and other resources required to reproduce our experiments.\footnote{\url{https://github.com/sdadas/vwsd}}

\section{System description}
\begin{figure*}
  \centering
  \includegraphics[scale=0.8]{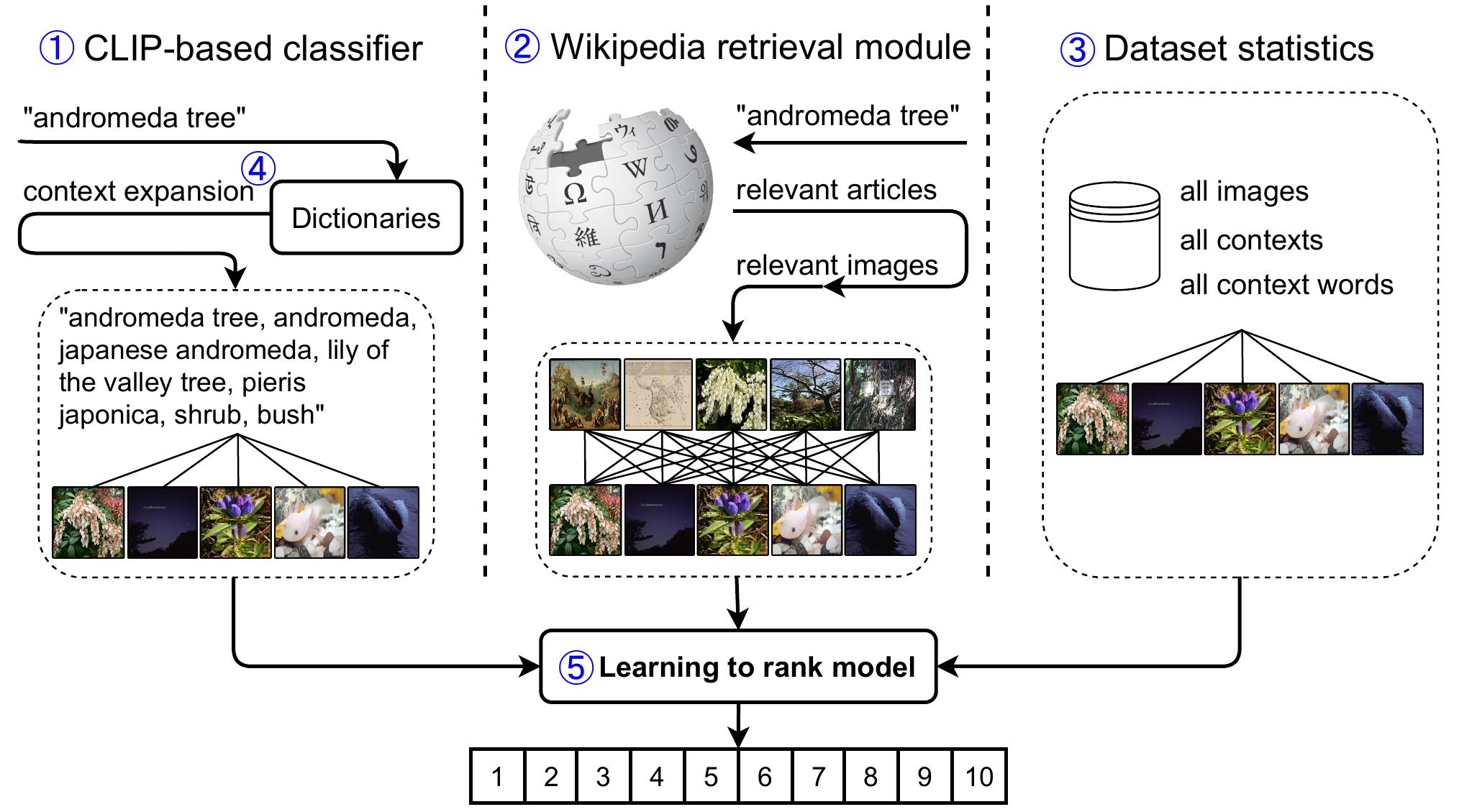}
  \caption{A diagram showing our visual word sense disambiguation system. Given the target word and its context, our method outputs a relevance ranking of candidate images. The ranking is produced by a fine-tuned learning to rank (LTR) model, which utilizes features extracted from the CLIP-based classifier, Wikipedia retrieval module, and global statistics calculated from the dataset.}
  \label{fig:model}
\end{figure*}

The proposed approach consists of several modules which together constitute a visual word sense disambiguation system. The three core components of our approach are: 1) a classifier based on CLIP image-text embeddings, 2) a Wikipedia retrieval module, 3) a learning to rank (LTR) model whose role is to generate the final ranking of images based on the features provided by the other modules. A high-level overview of the system is shown in Figure \ref{fig:model}.

\subsection{CLIP-based classifier \circled{1}}
CLIP (Contrastive Language-Image Pretraining) is a method for learning multimodal embeddings for language and vision by aligning the representations of images and their captions. The original CLIP models published by OpenAI \citep{radford2021learning} were trained on a dataset of 400 million image-text pairs. Recently, CLIP architectures trained on even larger datasets and with more parameters have been released by the LAION group, achieving state-of-the-art results on several image retrieval and zero-shot image classification tasks \citep{cherti2022reproducible}. Visual word sense disambiguation can be viewed as a text-to-image matching task, the type of problems for which CLIP is particularly effective. Therefore, we chose this model as the basis for our classifier.

We utilize CLIP in zero-shot mode, using a pre-trained checkpoint, to assign a score for each context-image pair from the data sample. Specifically, we compute vector representations of textual context $\bm{c}$ and image $\bm{x}$, and then calculate the similarity between these vectors using the following formula:
\begin{equation}
score(\bm{c},\bm{x})=sim(\bm{c},\bm{x}) - p(\bm{x})
\end{equation}
in which $sim(\bm{c},\bm{x})$ denotes a standard cosine similarity and $p(\bm{x})$ is a score penalty for the image $\bm{x}$. The penalty is calculated for each image as the mean similarity between that image and all the contexts in the dataset, normalized by the frequency of image occurrence. The rationale for using penalties is the observation that some images have high cosine similarity to many contexts, leading the model to incorrectly prefer them for the majority of samples in which they appear. The penalty lowers the similarity for these cases without affecting the results for the other images. We calculate it using the following formula:
\begin{equation}
p(\bm{x})=\left( \frac{1}{|C|}\sum_{\bm{c_{i}} \in C} sim(\bm{c_{i}},\bm{x}) \right) \cdot \frac{card(\bm{x})}{\max\limits_{\bm{x_{j}} \in X} card(\bm{x_{j}})}
\end{equation}
in which $C$ is the set of all contexts, $X$ is the set of all images, and $card(\bm{x})$ denotes the number of samples in which image $\bm{x}$ appears.

\subsubsection{Multilingual classification}
Publicly available CLIP models were trained on a set of English image captions, and are therefore not adapted for generating vector representations for texts in other languages. Consequently, the described method cannot be applied directly to Italian and Persian. However, it is possible to use transfer learning methods to train a multilingual or language-specific text encoder aligned with the representations generated by the original model. Such methods have been used in the past to create multilingual versions of CLIP \citep{reimers-2020-multilingual-sentence-bert,carlsson-etal-2022-cross}. The basic idea is to fine-tune a language model using bilingual or multilingual corpora. The original CLIP model generates a vector representation of the English text, while the language model produces a representation for the translation of that text. The difference between these vectors is then used to compute mean squared error (MSE) loss and the language model is optimized to approximate the representations generated by CLIP.

We employed this technique to train Italian and Persian text encoders, using OpenCLIP H/14 \citep{cherti2022reproducible}\footnote{\url{https://huggingface.co/laion/CLIP-ViT-H-14-laion2B-s32B-b79K}} as the teacher model and XLM-R large \citep{conneau2020unsupervised}\footnote{\url{https://huggingface.co/xlm-roberta-large}} as the student model. To train the encoders, we collected 10.5 million English images captions, which we then translated to Italian and Persian using publicly available neural machine translation models \citep{tiedemann-thottingal-2020-opus,khashabi-etal-2021-parsinlu}. The dataset for training was obtained from the following three sources:
\begin{itemize}[wide,labelwidth=0pt,labelindent=0pt,itemsep=0pt,topsep=8pt]
\item English subset of Wikipedia-based Image Text (WIT) dataset \citep{srinivasan2021wit}.
\item SBU Captions dataset \citep{ordonez2011im2text}.
\item A subset of 7 million English captions from Conceptual 12M \citep{changpinyo2021conceptual}.
\end{itemize}
For multilingual classification, we use the same scoring procedure as described in the previous section. The only difference is that we replace the original CLIP text encoder with our fine-tuned models.

\subsubsection{Context augmentation \circled{4}} 
One way to improve the performance of the described classifier is to expand the textual context with additional phrases associated with the actual meaning of the target word, which is expected to increase the similarity between the context and the correct image. We can do this with lexical databases by finding the sense of the target word and then extracting additional information from the definition of that sense. In our solution, we use multilingual resources available in Extended Open Multilingual WordNet \citep{bond2012survey,bond2013linking}. Specifically, we utilize the following lexical resources: 
\begin{itemize}[wide,labelwidth=0pt,labelindent=0pt,itemsep=0pt,topsep=8pt]
\item For English, we use Princeton WordNet database \citep{miller1995wordnet}.
\item For Italian, we use two lexical databases: one included in MultiWordNet \citep{Pianta:Bentivogli:Girardi:2002} and another one from EuroWordNet \citep{Toral:Bracale:Monachini:Soria:2010}.
\item For all three languages, we employ additional multilingual resources: Wiktionary and the Common Locale Data Repository (CLDR).
\end{itemize}

Our context expansion procedure works by appending alternative names extracted from a specific word sense, as well as from senses that are linked to it through hypernym, instance hypernym, member meronym, or substance meronym relations. For example, the context \emph{andromeda tree} is expanded to: \emph{andromeda tree, andromeda, japanese andromeda, lily of the valley tree, pieris japonica, shrub, bush}.

In order to find the correct sense, we retrieve a list of available senses of the target word from all lexical databases for a specific language and then compare the descriptions of these senses with the context word. The description is constructed from definitions, alternative names, and examples of use of a given sense, as well as senses linked to it by hypernym or instance hypernym relations. In our solution, we implemented two algorithms for matching sense and context:
\begin{itemize}[wide,labelwidth=0pt,labelindent=0pt,itemsep=0pt,topsep=8pt]
\item \textbf{Exact matching}, which involves finding exact occurrences of the context word in the sense description. The similarity between the context and the description is computed as the number of matched words divided by the total number of words in the description.
\item \textbf{Similarity matching}, involving the comparison of word vectors extracted from the word embedding model. In this method, we convert the context word and words from the sense description into their vector representations using multilingual FastText models \citep{grave2018learning}. The similarity between context and sense is calculated as the maximum cosine similarity between the representation of the context word and the representations of all the words from the description. 
\end{itemize}
We select the sense with the highest similarity to the context. For English, only exact matching is used. This method, however, has a low recall for languages other than English. For Italian and Persian we use exact matching first, and if no sense is found, we use similarity matching as a fallback method.

\subsubsection{Drawbacks of CLIP-based methods}
Although CLIP offers high zero-shot performance for the visual word sense disambiguation task, we also noticed certain problems in using this model that we could not fully eliminate. We share our observations below, which may provide suggestions for future research:
\begin{itemize}[wide,labelwidth=0pt,labelindent=0pt,itemsep=0pt,topsep=8pt]
\item The model is sensitive to images containing text. It also tends to assign high scores to images, which contain the target or context word. For example, for the context \emph{blue mood}, it assigned the highest similarity to an image showing just the word \emph{blue} on a blue background.
\item The model performs best with images, which directly show the object being described. However, it has trouble modeling more abstract relationships between textual context and image, especially when the context describes non-physical concepts such as emotions, actions, or events.
\item The model has a bias toward more commonly used word senses. As a result, in some cases even expanding the context with additional phrases directing the model to the correct prediction does not help, it still chooses the image relating to the more popular meaning of the target word.
\end{itemize}

\subsection{Wikipedia retrieval module \circled{2}}
Apart from the classifier described in the previous sections, our solution also includes a Wikipedia-based retrieval module, which returns an independent set of scores for each context-image pair. To apply this method, we first download publicly available Wikipedia dumps for the languages of interest, and then create BM25 \citep{robertson2009probabilistic} indexes with texts extracted from Wikipedia articles. For each document, we include a set of URLs to images attached to the article. We utilize WIT \citep{srinivasan2021wit} dataset to obtain a mapping between articles and images.

During inference, we use the following procedure to process the data sample, consisting of textual context and a set of images:
\begin{enumerate}[wide,labelwidth=0pt,labelindent=0pt,itemsep=0pt,topsep=8pt]
\item Full context is used to query the index. In response, we retrieve the top 10 articles sorted by their relevance to the query. If no relevant documents are found, we retry the search using only the target word as a query.
\item We download all the images attached to the retrieved articles. Next, we transform both the downloaded images and the images from the sample to their vector representations using the CLIP model.
\item We compare the sets of sample and article vectors using cosine similarity. The final score for each sample image is equal to the maximum of all similarities to the retrieved images.
\end{enumerate}

\subsection{Learning to rank \circled{5}}
The last element of our solution is the learning to rank model (LTR), which leverages the results returned by the other modules of the system to generate the final ranking of images. At the same time, it is the only component, which requires fine-tuning, as the modules described previously operate in zero-shot mode. Our approach is based on the LambdaMART algorithm \citep{burges2010ranknet}, which transforms the ranking problem into a pairwise classification task. It uses a loss function that compares the relative ordering of two items, rather than absolute scores. This allows the model to better capture the relative importance of different items in the ranking process. 

In our case, each data sample is represented by ten vectors consisting of numerical features, with each vector describing a comparison between the context and one of the sample images. We use the training set provided by the task organizers to optimize the model. The numerical features are computed from the outputs of the CLIP classifier and the Wikipedia retrieval module, as well as  calculated from the dataset statistics. In the case of scoring modules, the following features are extracted from each: the score assigned to the image, the average and maximum of the scores assigned to the other images from the same sample, the difference between the current score and the average, the difference between the current score and the maximum. We also include the penalty value $p(\bm{x})$ for the image, extracted from the CLIP-based classifier. As for other features, we include the following values in the model \circled{3}:
\begin{itemize}[wide,labelwidth=0pt,labelindent=0pt,itemsep=0pt,topsep=8pt]
\item Similarity values computed by CLIP between the image vector and the individual word vectors from the sample - separately for the target word and the context word.
\item Two frequency-related features, calculated as the logarithm of the number of occurrences of the image and the context word in the entire dataset.
\end{itemize}
Input features and hyperparameters of the LTR model are detailed in the Appendix.

\begin{table*}
\small
\centering
\setlength{\tabcolsep}{5pt}
\renewcommand{\arraystretch}{1.3}
\begin{tabular}{l|ccc|ccc|ccc|ccc}
\hline
\multirow{ 2}{*}{\textbf{System}} &  \multicolumn{3}{c|}{\textbf{Average}} & \multicolumn{3}{c|}{\textbf{English}} & \multicolumn{3}{c|}{\textbf{Italian}} & \multicolumn{3}{c}{\textbf{Persian}}\\
& \textbf{ACC} & \textbf{MRR} & \textbf{R} & \textbf{ACC} & \textbf{MRR} & \textbf{R} & \textbf{ACC} & \textbf{MRR} & \textbf{R} & \textbf{ACC} & \textbf{MRR} & \textbf{R} \\
\hline
Organizers' baseline & 37.20 & 54.39 & - & 60.48 & 73.87 & - & 22.62 & 42.61 & - & 28.50 & 46.70 & - \\
Best result for each language & & & & \textbf{84.02} & \textbf{89.55} & - & \textbf{84.26} & \textbf{89.05} & - & \textbf{64.00} & \textbf{74.39} & - \\
\hline
South China Normal University & \textbf{72.56} & \textbf{82.22} & \textbf{1} & 80.13 & 87.42 & 4 & 77.05 & 86.05 & 3 & 60.50 & 73.19 & 2 \\
Samsung Research China (Beijing) & 71.82 & 80.72 & 2 & \textbf{84.02} & \textbf{89.55} & \textbf{1} & 72.46 & 82.08 & 5 & 59.00 & 70.51 & 3 \\
\rowcolor[HTML]{eaecf0} Our system & 70.49 & 79.80 & 3 & 77.97 & 85.88 & 6 & 69.50 & 79.15 & 9 & \textbf{64.00} & \textbf{74.39} & \textbf{1} \\
\hline
\end{tabular}
\caption{\label{tab:results_main}
The performance of three top-rated teams in the visual word sense disambiguation task compared to the baseline solution provided by the organizers, as well as the highest scores achieved for each language, according to the official results. We show the average scores across all subtasks, as well as the results obtained on each language subtask. The table includes accuracy (ACC), mean reciprocal rank score (MRR), and the rank of each team (R). Bold values indicate the best result in a category.
}
\end{table*}

\section{Experiments and results}
This section contains a discussion of the official results of the visual word sense disambiguation task. We have also included a description of other variants of our system which were not used in the submitted solution. We conducted post-evaluation experiments using the gold labels provided by the organizers to analyze the results obtained by alternative versions of our approach.

\subsection{Official results}
The shared task consisted of three language subtasks, and the final ranking of the submitted solutions was based on the average of the results obtained in these subtasks. The primary metric used to evaluate the systems was accuracy, but the organizers also reported mean reciprocal rank (MRR) as an additional metric. 54 teams participated in the shared task. Our solution was ranked third in the main classification and won the Persian language subtask. The results of the top three ranked solutions, the official baseline, and the best results for each language subtask are shown in Table \ref{tab:results_main}.

The team which won the task achieved consistently high accuracy in all three languages, despite not winning on any of the subtasks. The other teams, including us, scored lower in one or more languages. The weak point of our solution was Italian, on which we ranked only 9th, with a difference of 14\% accuracy to the winning system. The performance of the best solutions on the Italian subtask turned out to be as high as on English, which was a surprise considering that no training data was available for this language.

\subsection{Post-evaluation results}
\begin{table}[h]
\small
\centering
\renewcommand{\arraystretch}{1.3}
\begin{tabular}{lcc}
\hline
\textbf{Method} & \textbf{ACC} & \textbf{MRR}\\
\hline
\multicolumn{3}{l}{\textbf{CLIP models}} \\
\hline
\multicolumn{3}{l}{\emph{OpenAI CLIP models}} \\
\hline
clip-vit-base-patch16 & 70.63 & 79.70 \\
clip-vit-base-patch32 & 71.92 & 80.56 \\
clip-vit-large-patch14 & 73.00 & 82.38 \\
\hline
\multicolumn{3}{l}{\emph{LAION CLIP models}} \\
\hline
CLIP-ViT-B-32-laion2B-s34B-b79K & 73.00 & 82.63 \\
CLIP-ViT-L-14-laion2B-s32B-b82K & 74.30 & 83.89 \\
\color{blue} CLIP-ViT-H-14-laion2B-s32B-b79K & \textbf{77.97} & \textbf{85.88} \\
CLIP-ViT-bigG-14-laion2B-39B-b160k & 76.89 & 85.48 \\
\hline
\multicolumn{3}{l}{\textbf{Context expansion methods}} \\
\hline
\color{blue} WordNet only & 77.97 & 85.88 \\
T5 only & 75.16 & 84.26 \\
WordNet + T5 & \textbf{78.83} & \textbf{86.26} \\
\hline
\end{tabular}
\caption{\label{tab:alternatives}
The performance of our system on the English subtask using alternative CLIP models or context expansion methods. Text in blue indicates the methods which were used in the submitted solution.
}
\end{table}

One of the most important components of our approach is the CLIP model, used for both text-to-image and image-to-image comparisons. In our solution, we employed \emph{OpenCLIP H/14} model published by LAION, which until recently was the largest CLIP variant available. In 2023, an even larger \emph{G/14} model was released. To study the impact of the selected CLIP version on the accuracy of the whole system, we tested our solution on available OpenAI and LAION models. The results of this experiment are shown in Table \ref{tab:alternatives}. Since we do not have multilingual versions of these models, the results presented are for the English subtask only. The conclusions of the experiment are consistent with results on other datasets found in the literature. LAION models achieve significantly higher accuracy than OpenAI models, and larger models outperform smaller ones. The only surprising finding is the weaker performance of the largest \emph{G/14} model. It is possible that other hyperparameters of our system would need to be readjusted in order to achieve the optimal performance for the largest model.

Another aspect of the system we examined is the context expansion method. In the submitted solution, we used a method based on WordNet and other lexical resources. While developing the system, we also explored an alternative technique using a sequence-to-sequence model. We employed the recently released Flan-T5 \citep{chung2022scaling} for this task. Our approach was to send the following prompt to the model: \emph{What is the meaning of [context]?} In response, the model would generate a definition of the given word sense, which we added to the context. The advantage of this approach is that it allows the context to be expanded for every sample, unlike dictionary-based methods which only expand the context with known definitions. The main disadvantage is the quality of the generated answers. In some cases, they were incorrect, which had a negative impact on the accuracy of the system. Therefore, we decided not to include this method in our solution. 

Table 2 shows the results of the three context expansion methods. We tested the performance of the approach based on the T5 model and lexical resources. We also tested a hybrid approach, in which we first try to expand the context using WordNet and other databases, and if that fails, we use the T5 model. As we expected, the standalone T5 model turned out to be worse than the lexical method. However, the hybrid approach managed to improve the accuracy of the English subtask over our original solution.

\subsection{Ablation study}
As part of the experiments, we performed an ablation study to better understand the impact of the various elements on the accuracy of our solution. The results are shown in Table \ref{tab:ablation}.

\begin{table}[h]
\small
\centering
\setlength{\tabcolsep}{3pt}
\renewcommand{\arraystretch}{1.3}
\begin{tabular}{l|cc|cc|cc}
\hline
\multirow{2}{*}{\textbf{Method}} &  \multicolumn{2}{c|}{\textbf{English}} & \multicolumn{2}{c|}{\textbf{Italian}} & \multicolumn{2}{c}{\textbf{Persian}}\\
& \textbf{ACC} & \textbf{MRR} & \textbf{ACC} & \textbf{MRR} & \textbf{ACC} & \textbf{MRR} \\
\hline
\color{blue} original & \textbf{77.97} & \textbf{85.88} & \textbf{69.50} & \textbf{79.15} & \textbf{64.00} & \textbf{74.39} \\
no penalties & 76.89 & 85.23 & 65.25 & 74.80 & 60.50 & 72.84 \\
no LTR & 75.80 & 84.61 & 68.19 & 77.76 & 59.00 & 70.50 \\
no expansion & 74.08 & 83.72 & 63.94 & 75.76 & 59.50 & 71.69 \\
no Wikipedia & 75.59 & 84.55 & 68.19 & 77.96 & 52.00 & 64.70 \\
CLIP only & 68.03 & 79.54 & 59.67 & 72.48 & 47.50 & 62.70 \\
\hline
\end{tabular}
\caption{\label{tab:ablation}
Ablation study of our system. The text in blue indicates the submitted solution.
}
\end{table}

The experiment involved performing an evaluation on a system in which a specific component was disabled. We disabled the following functionalities: penalties $p(x)$ used in the CLIP-based classifier (\emph{no penalties}), learning to rank model (\emph{no LTR}), context expansion (\emph{no expansion}), and Wikipedia retrieval module (\emph{no Wikipedia}). We also tested a version of the system stripped of all the above components, based only on the CLIP model (\emph{CLIP only}). In cases where LTR module is disabled, we instead use a simple heuristic to select the best matching image. We choose the image found by the Wikipedia retrieval module if the value assigned to that image is higher than 0.9, and the value assigned to the all other images is lower than 0.8. Otherwise, we select the highest rated image by the CLIP-based classifier.

As we can see, each of the components we proposed contributed to the performance of the final solution. The simplest version of the system, based only on the CLIP model, performs at least 10\% worse than the submitted solution. However, we can also observe that the effect of each functionality varies for different languages. For example, without Wikipedia, the English and Italian subtasks only lose approximately 2\% accuracy, while the Persian subtask scores 12\% lower. Context expansion is a feature, which has a significant performance impact on each of the three languages.

\section{Conclusion}
In this paper, we described our solution for the visual word sense disambiguation shared task. Our system was ranked third in the multilingual track and won in the Persian track, one of the three language-specific subtasks. In the publication, we demonstrated how to build a system, which incorporates different approaches to the problem: image-text embeddings, lexical resources, image and text retrieval. We showed that each of these components can improve the performance of the overall solution. We have also pointed out some drawbacks of our approach, which can be a starting point for creating better methods in the future.

\bibliography{anthology,custom}
\bibliographystyle{acl_natbib}

\clearpage
\appendix
\renewcommand{\arraystretch}{1.2}

\section{Hyperparameters and features}
\label{sec:appendix}

\begin{table}[h!]
\small
\centering
\begin{tabular}{ll}
\hline
\textbf{Hyperparam} & \textbf{Value}\\
\hline
Teacher model & CLIP-ViT-H-14-laion2B-s32B-b79K \\
Student model & XLM-Roberta-Large \\
Epochs & 3 \\
Batch size & 64 \\
Learning rate & 2e-5 \\
LR scheduler & Constant with warmup \\
Warmup steps & 2000 \\
Pooling & mean \\
\hline
\multicolumn{2}{l}{\textbf{Machine translation models used}} \\
\multicolumn{2}{l}{\textbf{Italian:} Helsinki-NLP/opus-mt-tc-big-en-it} \\
\multicolumn{2}{l}{\textbf{Persian:} persiannlp/mt5-large-parsinlu-translation\_en\_fa} \\
\hline
\end{tabular}
\caption{\label{tab:multilingual_params}
Hyperparameters for training Italian and Persian CLIP text encoders. We used a modified version of the script from Sentence-Transformers library: 
\url{https://www.sbert.net/examples/training/multilingual/README.html\#training}
}
\end{table}

\begin{table}[h!]
\small
\centering
\begin{tabular}{ll}
\hline
\textbf{Hyperparam} & \textbf{Value}\\
\hline
Tree learning method & gpu\_hist \\
Loss function & pairwise \\
Number of trees & 110 \\
Max tree depth & 6 \\
Learning rate & 0.1 \\
Subsample columns & 0.9 \\
Subsample data & 0.75 \\
\hline
\end{tabular}
\caption{\label{tab:ranker_params}
Hyperparameters for training learning to rank model. We used XGBRanker class from XGBoost library.
}
\end{table}

\begin{table}[h!]
\small
\centering
\begin{tabular}{l}
\hline
\textbf{Features from CLIP-based classifier}\\
\hline
A. Image score \\
B. Maximum of the scores of other images \\
C. Average of the scores of other images \\
D. Difference between A and B \\
E. Difference between A and C \\
F. Penalty score $p(x)$ for the image \\
\hline
\textbf{Features from Wikipedia retrieval module}\\
\hline
G. Image score \\
H. Maximum of the scores of other images \\
I. Average of the scores of other images \\
J. Difference between G and H \\
K. Difference between G and I \\
\hline
\textbf{Additional features}\\
\hline
L. Similarity between the image and the target word \\
M. Similarity between the image and the context word \\
N. $\log_{10}(card(x))$ for the image $x$ \\
O. $\log_{10}(card(w))$ for the context word $w$ \\
\hline
\end{tabular}
\caption{\label{tab:ranker_features}
A list of features used by the LTR model.
}
\end{table}

\end{document}